\documentclass{esannV2}
\usepackage{graphicx}
\usepackage[latin1]{inputenc}
\usepackage{amssymb,amsmath,array}
\usepackage[dvipsnames]{xcolor}
\usepackage{url}

\begin{document}

\title{Sensivity of LLMs' Explanations to the Training Randomness:\,Context,\,Class\,\&\,Task\,Dependencies}

\author{Romain Loncour and J\'{e}r\'{e}mie Bogaert and Fran\c{c}ois-Xavier Standaert
\vspace{.3cm}\\
UCLouvain, Louvain-la-Neuve, Belgium.
}

\maketitle

\begin{abstract} Transformer models are now a cornerstone in natural language processing. Yet,  explaining their decisions remains a challenge. It was shown recently that the same model trained on the same data with a different randomness can lead to very different explanations. In this paper, we investigate how the (syntactic) context, the classes to be learned and the tasks influence this explanations' sensitivity to randomness. We show that they all have statistically significant impact: smallest for the (syntactic) context, medium for the classes and largest for the tasks.
\end{abstract}

\section{Introduction}

Transformer models~\cite{DBLP:conf/nips/VaswaniSPUJGKP17,DBLP:conf/naacl/DevlinCLT19} are increasingly used in Natural Language Processing (NLP)~\cite{DBLP:journals/information/PatwardhanMS23}. However, explaining their decisions has been recurrently shown to be difficult: various methods have been proposed~\cite{DBLP:journals/information/PatwardhanMS23}, offering different trade-offs between a varying set of desirable properties~\cite{DBLP:journals/coling/LyuAC24}. Among these properties, the faithfulness (i.e., the extent to which the explanation reflects the model's behavior) and the plausibility (i.e., the extent to which the explanation is understandable and convincing to a human being) usually come first.

\smallskip

While explaining one instance of a transformer model is already an important  challenge, it was recently shown that the same model trained on the same data but with a different randomness can lead to significantly different explanations~\cite{DBLP:conf/pkdd/MullerTBJBW23,TAL}.\footnote{~An English translation of~\cite{TAL} is available on ArXiv~\cite{DBLP:journals/corr/abs-2410-05085}.} 
This raises a need to investigate
the impact and the dependencies of the training randomness on the explanations,  considering the distribution of the explanations rather than single instances, as dominantly analyzed so far. 

\smallskip

In this paper, we contribute to this challenge by assessing three natural dependencies.
We first study the impact that the (syntactic) context has on the explanations' sensitivity to the training randomness. We then evaluate the influence of the class to be learned. More precisely, we show that the explanations of classes characterized by the presence or absence of discriminant markers have a different sensitivity to the training randomness. We finally discuss the task-dependency of this sensitivity to the training randomness. 
Our results show that all these factors have a statistically significant impact: smallest for the (syntactic) context, medium for the classes and largest for the tasks. We conclude by discussing the interesting open questions raised by these findings. 

\section{Experiment design}

\subsection{Methods}

We next evaluate the explanations' sensitivity to the training randomness for different
datasets of the same size.
To classify English pieces of text, we use the RoBERTa-base model \cite{DBLP:journals/corr/abs-1907-11692}. When working with the French language, we use the CamemBERT-base model \cite{DBLP:conf/acl/MartinMSDRCSS20}. In all cases, we fine-tuned 200 models using the same hyper-parameters (which are the default ones, aside from the learning rate of $2 \times 10^{-5}$, the batch size of 16 and the number of epoch of 1), and a different random seed each time. The seed controls the order of the training dataset, the deactivated neurons based on the dropout and the initialization of the transformer models' classification head. We first selected a subset of $m$ equivalent models, whose accuracy does not significantly differ on our test set. 
We then selected so-called compatible texts (i.e., texts on which all models predict the same label) from the test sets of our datasets.
We finally produced explanations for the corresponding model predictions using Layer-wise Relevance Propagation (LRP) explanations~\cite{bach2015pixel,DBLP:conf/cvpr/CheferGW21}. It is a deterministic explanation method which has been shown to provide a reasonable trade-off between plausibility and faithfulness \cite{DBLP:journals/coling/LyuAC24}. Since LRP explanations provide one explanation value per token, we therefore obtain $m$ vectors of $n$ values when explaining an $n$-word text.

\subsection{Metric}

\begin{figure}[h!]
    \centering
    \includegraphics[width=0.8\linewidth]{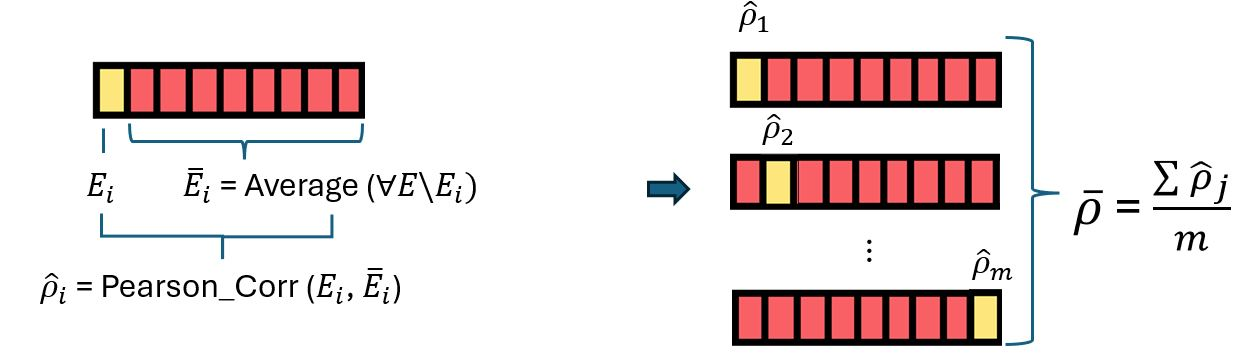}
    \caption{MCWME estimation with Pearson's correlation coefficient.}
    \label{fig:MCWME}
\end{figure}

As an explanation stability metric, we use the Mean Correlation With Mean Explanation (MCWME) metric~\cite{DBLP:conf/xai/BogaertDS25}. 
This metric allows computing the explanation stability across $m$ explanations for any 
$n$-word text. 
Figure \ref{fig:MCWME} illustrates how MCWME can be estimated. Essentially, it works
by computing a mean explanation with a part of the dataset and testing how correlated the explanations of the other (equivalent) models are with this mean explanation. 
We use leave-one-out cross-validation to obtain the most accurate estimates with the available data.  That is, $m$ correlations are computed 
between an explanation and the average explanation (leaving aside that one explanation)
and then averaged.

\section{Impact of the syntactic context}

Our first experiment investigates the impact that shuffling the sentences' words in a dataset has on the explanations' stability. To run this experiment, we used two datasets: the first one is composed of a set of $10~000$ different 10-word long sentences, split in two classes, where only one word differs. The second one is a shuffled version.
To avoid incoherent sentences when switching words, we make use of first names. The first class always contains the first name ``John'', while it is replaced by the first name ``James'' in the second one. The second dataset is composed of the exact same sentences, but the words composing them are shuffled. It follows that these two datasets are composed of the exact same word distribution, but that the order of those words varies from one to the other. We then selected two (i.e., one for each training dataset) subsets of $m=100$ equivalents models.
All models reach 100\% test accuracy on their respective task. We finally computed the
MCWME metric for 80 different texts.

\smallskip

\begin{figure}[h!]
    \centering
    \includegraphics[width=\linewidth]{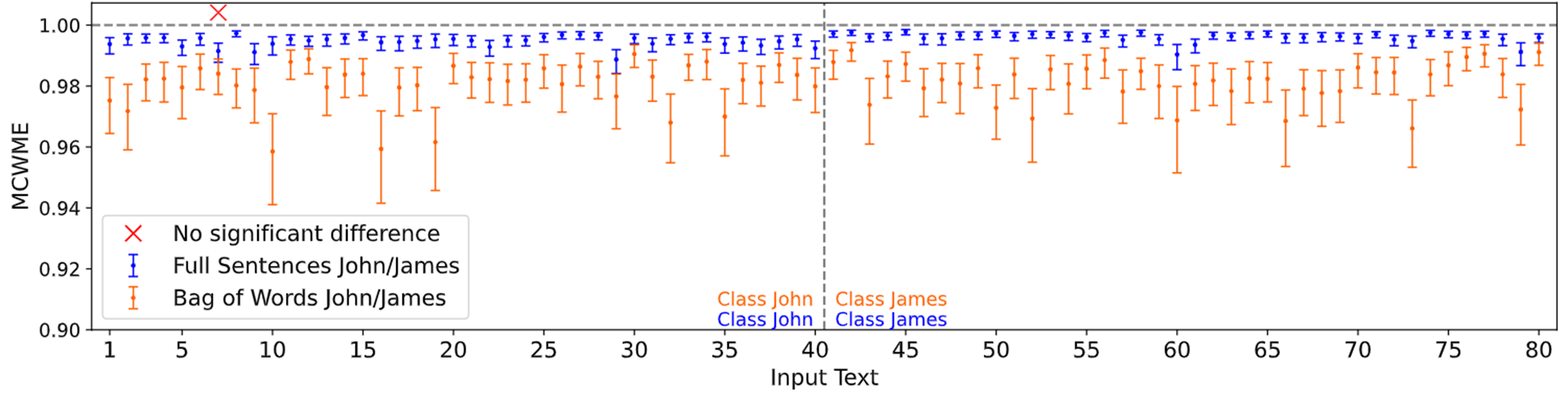}
    \caption{MCWME comparison plot between shuffled and non-shuffled sentences. A red cross on top of the plot highlights a non significant difference}
    \label{fig:fsjj-vs-bowjj}
\end{figure}

Our results are summarized in 
Figure~\ref{fig:fsjj-vs-bowjj}. Each \textit{x}-coordinate corresponds to an MCWME and its confidence interval for a given text. Since each (ordered) text in the first dataset has a corresponding shuffled text in the second dataset, we can pair them together and proceed to multiple comparisons on the same plot. The results for the two classes are separated by a dashed line. 

\begin{figure}[h!]
    \centering
    \includegraphics[width=1\linewidth]{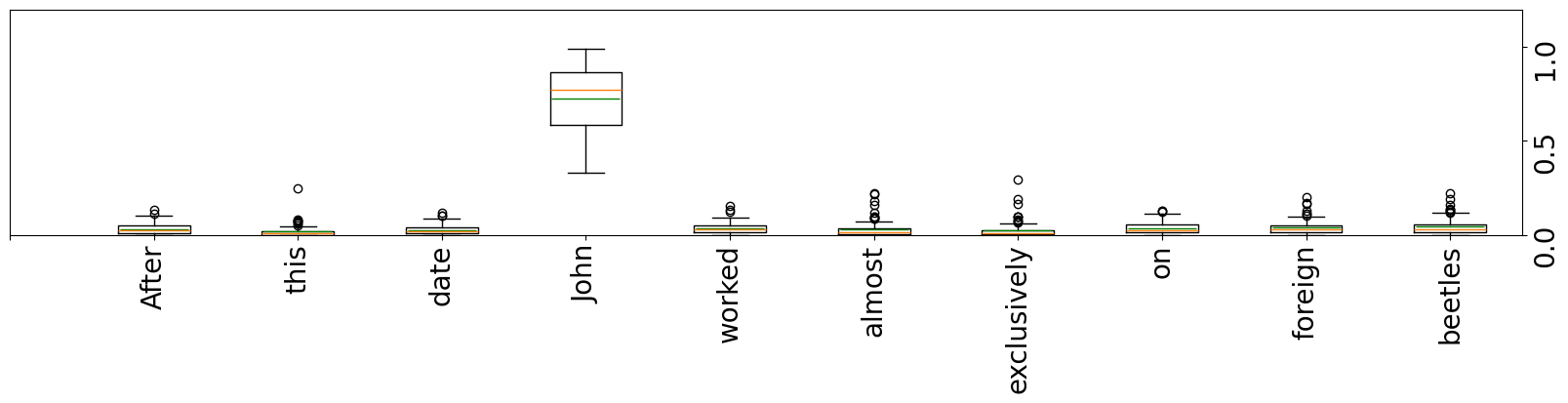}
    \caption{Explanations' boxplot for a text containing the first name ``John''.}
    \label{fig:bp_jj_john}
\end{figure}

It shows
close to perfect explanation stability when the words are not shuffled (blue results).
Figure \ref{fig:bp_jj_john} further confirms that the explanations only provide a high-relevance value to the discriminant word of the task. This strengthens our confidence in LRP's capacities to provide reasonably faithful explanations.

\smallskip

Regarding our research question, we observe that a vast majority of MCWME comparisons between one text and its shuffled  counterpart (i.e., apart for a single red cross)
show significant differences. We conclude that though explanations are quite stable for such a simple task, shuffling the words during the fine-tuning leads to a slightly lower explanations' stability (i.e., a higher explanations' sensitivity to the training randomness).
This is presumably due to the transformer models characterizing small relations between words that should be set to zero and are nevertheless reported by the attention mechanism of LRP. 
It suggests the use of simple(r) models whenever they do not imply accuracy losses.

\section{Class dependency: absence of discriminant words} 

One of the fundamental characteristic of LRP explanations is that they can only provide a relevance value to each word present in the text. In this second experiment, we investigate the explanations of classes that are characterized by the absence of discriminant words which, by  design, cannot be highlighted in the text. We use two datasets for this purpose: the first one is the same as in the previous section. The second one is built by keeping the class corresponding to the first name ``John'' and by replacing the first name ``James'' by a random word in the second class (which may create coherence issues within the sentences).\footnote{~We also investigated the removal the first name in the second class, but the size of the text then becomes discriminant while
raises the same coherence issues as the proposal above.} The task can still be learned with perfect accuracy, but the label is now determined solely by the presence or absence of the first name ``John''. 

\begin{figure}[h!]
    \centering
    \includegraphics[width=\linewidth]{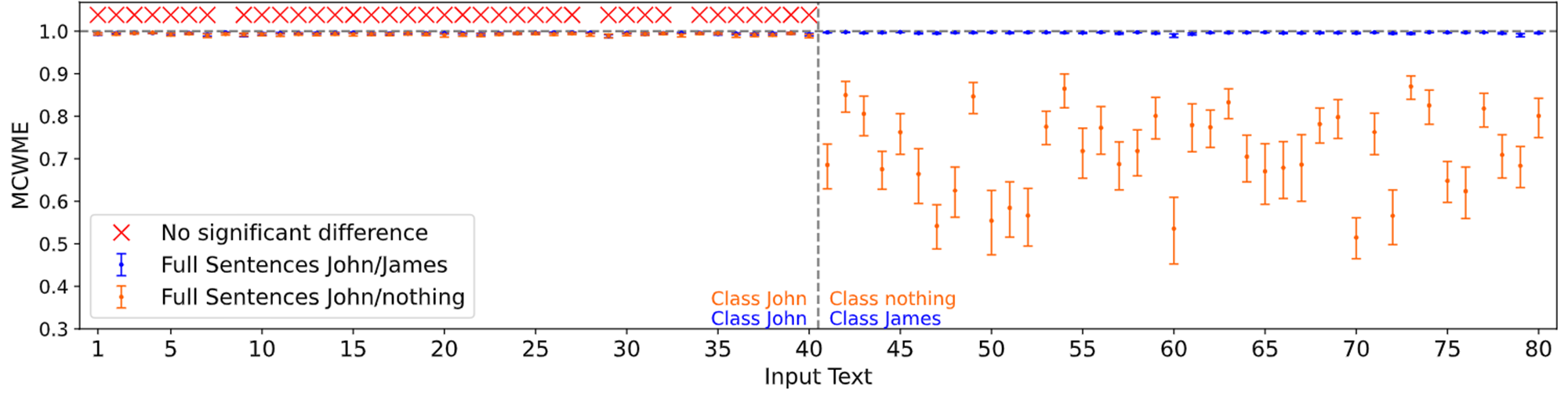}
    \caption{MCWME comparison plot. Texts of the left class always contain the first name ``John''. Texts on the right contain either  ``James'' 
    or no first name.}
    \label{fig:fsjj-vs-fsjn}
\end{figure}

\begin{figure}
    \centering
    \includegraphics[width=1\linewidth]{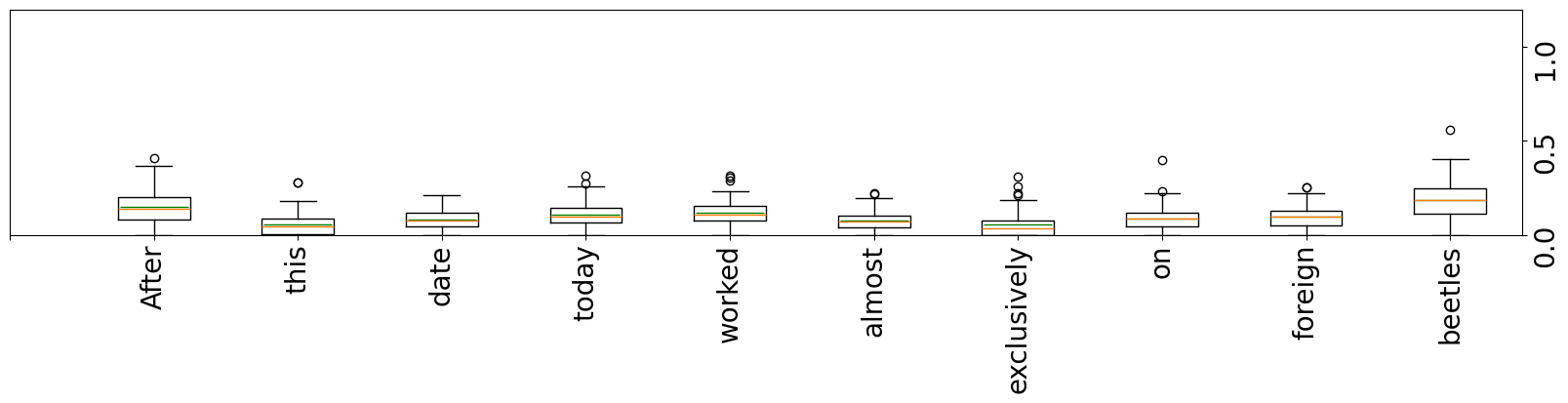}\vspace*{-0.2cm}
    \caption{Explanations' boxplot for a text where ``John'' is replaced by ``today''.}
    \label{fig:bp_jn}\vspace*{-0.3cm}
\end{figure}

This time, we observe that the MCWME is significantly lower for the class without any discriminant marker. It confirms that the explanations' sensitivity to the training randomness can be class-specific. Interestingly, we observe MCWME values around $0.7$, while random explanations would provide values around 0. These values presumably arise because all words may not be considered as equally irrelevant in the explanations. For example, despite the average explanation being ``flatter'', the words with the highest average relevance seem to be at the beginning and end of the sentences, and around the replaced word. This is illustrated in Figure \ref{fig:bp_jn} and left as a 
scope for further research (e.g., it could be interesting to check whether correlation vanishes with larger texts).

\section{Real world use-case : ArXiv vs InfOpinions} 
\vspace*{-0.1cm}
We finally compare the explanation stability across two different tasks. The first one consists in categorizing paper abstracts related to astro-physics or mathematics. To train our models, we use the ArXiv dataset from which we kept 5000 abstracts with the Astro-ph.GA and 5000 with the Math.NT tags.\footnote{~\url{https://www.kaggle.com/datasets/Cornell-University/arxiv}} The texts are 148.43 tokens long on average. The second task consists in classifying press articles into the information and opinion categories. We use the InfOpinion dataset \cite{bogaert2023} for this purpose. The texts are 338.6 tokens long on average. Models trained on the Arxiv dataset reach 99.8\% accuracy on the test set, while models trained on InfOpinion reach around 96\%. Since we have no way to pair the texts from the InfOpinion and ArXiv datasets together, we pair them randomly.

\begin{figure}[h!]
    \centering
    \includegraphics[width=\linewidth]{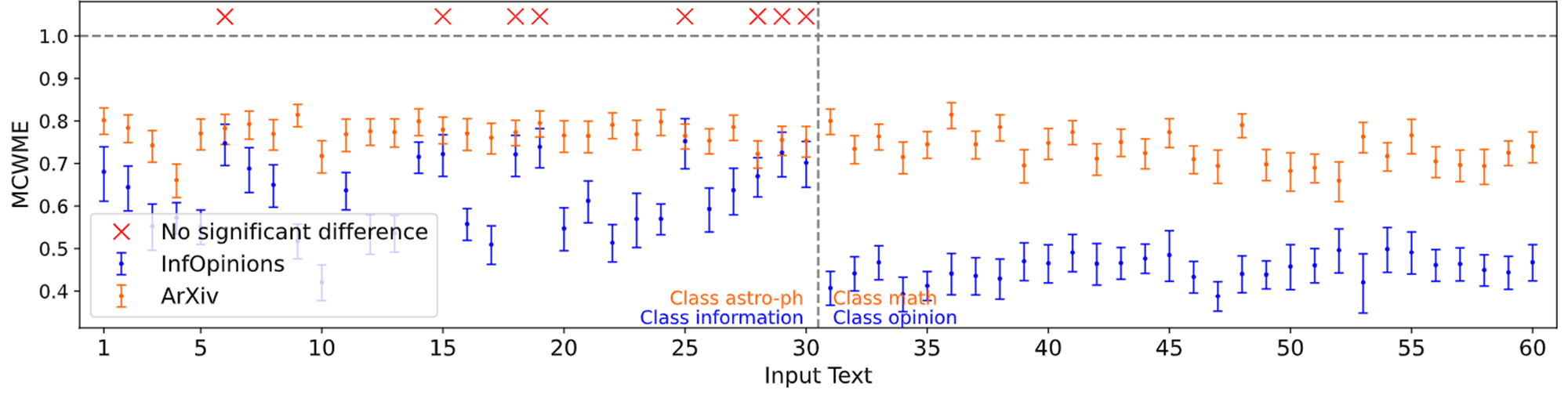}
    \caption{InfOpinions vs. ArXiv}
    \label{fig:infopinions-vs-arxiv}
\end{figure}

Figure \ref{fig:infopinions-vs-arxiv} shows a MCWME comparison plot over the two tasks. We first observe that the explanation stability is different between the two classes of the InfOpinion dataset (in blue), confirming that the trend put forward in the last section can also be observed when working with real-world data. 

\smallskip

We also observe a significant difference in the explanation stability on both tasks. We posit that this gap comes from the more discriminant vocabulary between the classes in the ArXiv dataset than in the one in the InfOpinion dataset, where a more in depth understanding of the relations between the different words is needed. This hypothesis is also made plausible by the fact that the models perform better on the ArXiv dataset than on the InfOpinion one, reflecting that one task is easier than the other. Other factors can nevertheless play a role in this difference as well, such as the length or the language of the texts. 

\section{Conclusion}

The dependencies we put forward suggest that characterizing the explanations' sensitivity to the training randomness of large language models could be a useful addition to existing explainability frameworks. They raise as main open question their impact on the  plausibility and their link with the faithfulness of the explanations. On the plausibility side, interpreting a distribution of explanations asks to process more information than interpreting a single explanation. On the faithfulness side, it is questionable whether more complex explanation methods could lead to a reduced dependency on the training randomness of~complex~models.

\medskip

\noindent\textbf{Acknowledgments.} Fran\c{c}ois-Xavier Standaert is a research director of the Belgian Fund for Scientific Research (FNRS-F.R.S.). This work has been supported
by the Service Public de Wallonie Recherche, grant 2010235-ARIAC.

\begin{footnotesize}

\end{footnotesize}
\end{document}